\documentclass[letterpaper]{article} % DO NOT CHANGE THIS
\usepackage{aaai2026}
\nocopyright
\usepackage{amsfonts}
\usepackage{amsmath}
\usepackage{booktabs}
\usepackage{multirow}
\usepackage{bm}
\usepackage{times}  % DO NOT CHANGE THIS
\usepackage{helvet}  % DO NOT CHANGE THIS
\usepackage{courier}  % DO NOT CHANGE THIS
\usepackage[hyphens]{url}  % DO NOT CHANGE THIS
\usepackage{graphicx} % DO NOT CHANGE THIS
\usepackage{natbib}  % DO NOT CHANGE THIS AND DO NOT ADD ANY OPTIONS TO IT
\usepackage{caption} % DO NOT CHANGE THIS AND DO NOT ADD ANY OPTIONS TO IT
\usepackage{algpseudocode}
\usepackage{algorithm}

\usepackage{newfloat}
\usepackage{listings}
\DeclareCaptionStyle{ruled}{labelfont=normalfont,labelsep=colon,strut=off} % DO NOT CHANGE THIS
\floatstyle{ruled}
\newfloat{listing}{tb}{lst}{}
\floatname{listing}{Listing}
\title{Shapelets-Enriched Selective Forecasting using \\Time Series Foundation Models}
\author {
    Shivani Tomar\textsuperscript{\rm 1},
    Seshu Tirupathi\textsuperscript{\rm 2},
    Elizabeth Daly\textsuperscript{\rm 2},
    Ivana Dusparic\textsuperscript{\rm 1}
}
\affiliations {
    \textsuperscript{\rm 1}Trinity College Dublin\\
    \textsuperscript{\rm 2}IBM Research, Dublin \\
    tomars@tcd.ie, seshutir@ie.ibm.com, elizabeth.daly@ie.ibm.com, ivana.dusparic@tcd.ie
}

\begin{document}

\maketitle

\begin{abstract}
Time series foundation models have recently gained a lot of attention due to their ability to model complex time series data encompassing different domains including traffic, energy, and weather. Although they exhibit strong average zero-shot performance on forecasting tasks, their predictions on certain critical regions of the data are not always reliable, limiting their usability in real-world applications, especially when data exhibits unique trends. In this paper, we propose a selective forecasting framework to identify these critical segments of time series using shapelets. We learn shapelets using shift-invariant dictionary learning on the validation split of the target domain dataset. Utilizing distance-based similarity to these shapelets, we facilitate the user to selectively discard unreliable predictions and be informed of the model's realistic capabilities.
Empirical results on diverse benchmark time series datasets demonstrate that our approach leveraging both zero-shot and full-shot fine-tuned models reduces the overall error by an average of 22.17\% for zero-shot and 22.62\% for full-shot fine-tuned model. Furthermore, our approach using zero-shot and full-shot fine-tuned models, also outperforms its random selection counterparts by up to 21.41\% and 21.43\% on one of the datasets.
\end{abstract}

\section{Introduction}

Time series forecasting has seen tremendous advancements using general purpose pre-trained time series foundation models (TSFMs) such as Chronos \cite{b7}, Tiny Time Mixers (TTM) \cite{b2}, Moirai \cite{b8}, TimesFM \cite{b9}, LLMTime \cite{b10}, and GPT4TS \cite{b11}. These models exhibit strong benchmark performances in zero-shot forecasting with further improvements gained with fine-tuning. However, these models report their performance using average error but often fail to provide reliable predictions in specific segments characterized by patterns of significant or abrupt changes \cite{b28}. This affects the reliability of such models in real-life applications. The technique of selective forecasting was introduced a long time ago \cite{b20, b21}. It involves learning   predictive models that are allowed to abstain from predicting if they are not confident in their predictions. There are two ways of building selective predictors : First, add a selection mechanism to a trained prediction model, and second, jointly train the prediction model along with the selection function \cite{b1}. In order to address the shortcoming associated with unreliable predictions, in this paper, we adopt the first method where we use shapelet-based distance similarity as the selection criteria for discarding model predictions. Shapelets were introduced by \cite{b4} as a time series primitive in the area of time series data mining particularly suited to the downstream task of time series classification, clustering and anomaly detection. Shapelets are time series subsequences that are maximally representative of a class. They are often used in time series classification problems due to their inherent interpretability. We use these shapelets to capture the local patterns in time series segments where the TSFM provides potentially unreliable predictions that are prone to high error. They serve as the basis of the sample selection mechanism used as part of our approach for selective forecasting. We employ a light-weight time series foundation model, Tiny Time Mixer (TTM) \cite{b2} as the pre-trained prediction model leveraging its robust zero-shot forecasting performance with minimal computational overhead. Our approach allows the user to reduce the overall error by discarding unreliable predictions.

The main contributions of this paper are as follows:

\begin{itemize}
    \item Proposes a selective forecasting framework, utilizing time series shapelets to identify the occurrence of specific patterns in data which lead to potentially high prediction errors on the target domain datasets. 
    \item We introduce a sample selection module using distance-based similarity with learned shapelets to perform selective forecasting.
    \item Extensive experiments using TTM, a light-weight time series foundation model on benchmark datasets prove the efficacy of our framework to improve the overall forecasting performance on the target domain ensuring flexibility to discard unreliable predictions. Our approach leads to significant reduction in mean squared error (MSE), averaging 22.17\% for zero-shot and 22.62\% for full-shot fine-tuned models. It also outperforms random selection baseline using zero-shot and full-shot fine-tuned model by significant margins of 21.41\% and 21.43\% respectively on one of the datasets.
\end{itemize}

\section{Related Work}

Recent efforts in the field of time series forecasting using deep learning \cite{b12, b13, b14, b15} have led to improved long-term forecasting accuracy with reduced latency. Furthermore, TSFMs have been gaining popularity for their extraordinary zero-shot performance on forecasting time series. There is ongoing research to further improve their domain-specific forecasting capabilities while still retaining the generalized representations learned during pre-training phase. That led to the adoption of parameter efficient fine-tuning (PEFT) techniques \cite{b23} focused on updating only a fraction of weights to learn domain-specific features while keeping the remaining weights frozen. LoRA \cite{b22} has emerged as an effective PEFT technique resulting in superior fine-tuned performance especially in the case of scarce datasets. Another work \cite{b24} aims to improve the naive fine-tuning process by analyzing it through the causal perspective. They explicitly model multiple scales encountered in time-series target domain task during fine-tuning to better capture the temporal patterns. Another interesting aspect captured by Zhao et al. in \cite{b25} reveals the sparsity in hidden states and redundancy in parameters inherent in TSFMs. They use these substructures to prune the TSFM before fine-tuning leading to further performance improvements during inference time. Tomar et al. \cite{b27} developed a novel framework named AT4TS, for automated fine-tuning of TSFMs combining parameter efficient fine-tuning methods with state-of-the-art hyperparameter optimization techniques leading to significant improvement for out-of-domain datasets. While improved results have been demonstrated by the above approaches, current methods do not explore the application of selective forecasting in relation to TSFMs.

In this work, we focus on these small percentage of time series segments to reason and improve the model's overall performance. A lot of research has been conducted in the field of selective prediction also known as reject option techniques mainly using algorithms like SVMs, boosting and nearest neighbours \cite{b16, b17}. Recently, these approaches have been extended to deep learning algorithms \cite{b18}. SelectiveNet \cite{b18} was the first work which trains to optimize for both the classification loss and the reject function during the training process. However, in case of TSFMs which are already pre-trained, simultaneous training of a selective reject function is not feasible. To this effect, we use a shapelet module to selectively reject predictions for samples during inference time which are similar to the shapelets learned on the validation split. In the past decade, Shapelets \cite{b4} have garnered increasing attention in the context of time series classification. Shapelets are discriminative subsequences of time series that maximally distinguish among different classes by discovering local shape features inherent in each class. This makes them suited to downstream tasks of classification, clustering and anomaly detection. However, their use in the context of forecasting remains underexplored.

%%%%%%%%%%%%%%%%%%%%%%%%%%%%%%%%%%%%%%%%%%%%%%
\begin{figure*}[t]
\centering
\includegraphics[width=0.8\textwidth]{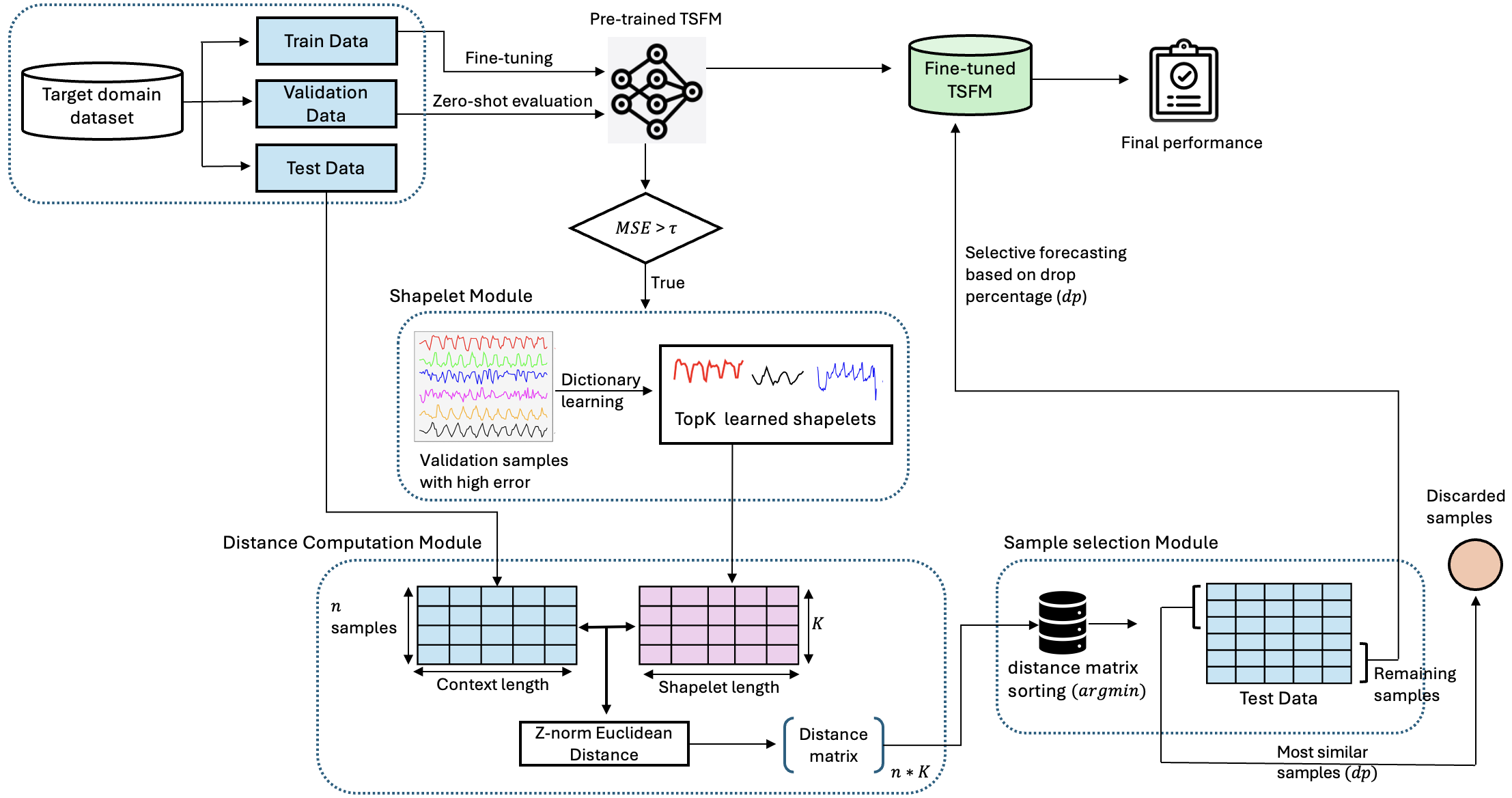}
\caption{Workflow diagram showing the selective prediction  framework guided by shapelets.}
\label{fig:arch_diag}
\end{figure*}
%%%%%%%%%%%%%%%%%%%%%%%%%%%%%%%%%%%%%%%%%%%%%%

\section{Proposed Methodology}

In this section, we describe the detailed methodology adopted in this paper for selective forecasting using TSFMs guided by shapelets. As shown in Figure~\ref{fig:arch_diag}, we split the target domain dataset into train, validation and test split and use a pre-trained TTM for both zero-shot predictions and fine-tuning. Algorithm~\ref{alg:algorithm} outlines the steps involved in our approach.

\textbf{Zero-shot performance on held-out Validation set} : We first use a pre-trained TSFM, in our case, TTM, to make zero-shot predictions on the validation split of the dataset, $X_{\text{val}}$. We formally define the task of time series forecasting below. Let $\bm{X} \in \mathbb{R}^{c \times sl}$ be a multivariate time series of length $sl$ and number of channels $c$. The forecasting task involves predicting the future values $\bm{Y} \in \mathbb{R}^{c' \times fl}$ given the history/context $\bm{X}$. Here, $fl$ denotes the forecast length/horizon, and $c'$ denotes number of forecast channels, where $c' \le c$. In our case, we set the number of channels, $c$ to 1 for univariate forecasting. The predictions from the model are denoted by $\hat{\bm{Y}} \in \mathbb{R}^{c' \times fl}$. 

We use a pre-trained time series foundation model, TTM, and make predictions $\hat{\bm{Y}}$ on the held-out validation split of the unseen target domain dataset. We use $MSE$, as the evaluation metric used by the original paper~\cite{b2}. Based on the average MSE obtained on the entire $X_{\text{val}}$, we set the error threshold $\tau$ to identify the segments of the time series where the zero-shot performance of the model is unreliable. We set $ \tau = (err + \delta * std)$ using $\delta$ as the multiplier to vary the threshold as shown in Step 4 of Algorithm~\ref{alg:algorithm}. This step involves analyzing the performance of the pre-trained TSFM on the unseen target domain downstream task without any kind of adaptive fine-tuning. Therefore, it enables us to establish the average baseline performance on the target domain dataset.

\textbf{Full-shot Fine-tuning TSFM on Train set} : As shown in step 1 of Algorithm~\ref{alg:algorithm}, we use the train split of the target data to fine-tune the pre-trained TSFM. We later use this fine-tuned model (${M}^*$) as the baseline to compare against our approach.

\textbf{Shapelet module} : This module is central to our framework as it defines the criteria to filter the test samples for making selective predictions using zero-shot/full-shot fine-tuned TSFM. We utilize the shift invariant dictionary learning approach \cite{b3} to learn a set of basis elements from the validation set. These basis elements serve as shapelets. We would like to clarify that in the time series literature, shapelets \cite{b4, b5, b6} are defined as time series subsequences in the training data that are useful in discriminating among classes. However, in this work, we do not use shapelets in the context of identification of time series classes but purely to uncover representative patterns in time series data. We first identify samples of time series sequences in the validation set that correspond to high zero-shot prediction error. The error threshold is decided based on the average zero-shot error on the validation split of the target domain dataset. Shapelets are then learned from these curated high error-causing validation samples. The number(\textit{K}), length(\textit{q}) of shapelets and sparsity regularization parameter $\lambda$ are selected using grid search over 3-fold cross-validation on the validation split. The property of shift invariance makes the application of dictionary learning useful for extracting representative patterns even in the absence of class labels. For a time series $\left\{ \mathbf{x}_i \in \mathbb{R}^p \right\}_{i=1}^n$ with $n$ input data points and dimension $p$, authors in \cite{b3}, relaxed the length constraint of $K$ basis vectors such that length of each basis or dictionary element \( q \leq p \). Note that $K$ basis atoms comprise the dictionary $D = [d_{1}, ...d_{k}] \in \mathbb{R}^{q \times K}$ and for each $x_{i}$, the sparse coding $\alpha_{i} \in \mathbb{R}^K$ is learned along with the dictionary.  In order to achieve the shift invariance property, a new variable $t_{ik}$ is introduced, which specifies the offset where $d_{k}$ can be overlaid to match $x_{i}$ such that $t_{ik}$ can only take integer values in the range $[0, ..p-q]$. 

Therefore, problem of learning both the sparse coding and dictionary elements or shapelets is given by 

\begin{align}
\arg \min_{\substack{\mathbf{D} \in \mathbb{R}^{q \times K} \\ \boldsymbol{\alpha}_i \in \mathbb{R}^K \\ t_{ik} \in [0, p-q]}} 
&\frac{1}{2} \sum_{i=1}^{n} \left\| \mathbf{x}_i - \sum_{k=1}^{K} \boldsymbol{\alpha}_{ik} T(\mathbf{d}_k, t_{ik}) \notag \right\|_2^2 + \\  \lambda \sum_{i=1}^{n} \|\boldsymbol{\alpha}_i\|_1 \\
&\text{s.t.} \|\mathbf{d}_k\|^2 \leq c, \quad \text{for } k = 1, \ldots, K 
\end{align}

Similar to the classical dictionary solving approach, the above problem is also solved using gradient descent approach.

Based on the number of shapelets ($K$) learned, we further reduce the dictionary size to avoid redundant/self-similar shapelets by selecting $TopK$ shapelets by using the ranking obtained from sparse coding $\alpha$ matrix.

\textbf{Distance Computation Module}: Post shapelet learning module described above, we take the \textit{TopK} shapelets which basically represent the patterns that coincide with higher than average prediction error on the validation split. We then use z-normalized euclidean distance-based similarity matching on the test data samples which is implemented as $\text{DISTANCE\_COMP}$ procedure in Algorithm~\ref{alg:algorithm}. This yields a distance matrix ($D_{mat}$) as shown in Step 11.

\textbf{Sample Selection Module}: The distance matrix ($D_{mat}$) computed in the above module is then sorted in ascending order to retrieve the indices of test samples that closely match the shapelets. The $\text{DISCARD}$ procedure mentioned in Step 14 is used to drop the test samples and their corresponding prediction error is marked as 0. The drop percentage $dp$ is a user-defined threshold to specify the ratio of predictions that are allowed to be discarded from the final set of predictions made by the TSFM on the test set to ensure adequate coverage.

\begin{algorithm}[tb]
\caption{Selective Forecasting using TSFM guided by Shapelets }
\label{alg:algorithm}
\textbf{Input:} Pre-trained TSFM, $TTM$, target dataset $X$ split into $X_{\text{train}}$, $X_{\text{val}}$, and $X_{\text{test}}$,  user-defined drop percentage $\textit{dp}$, multiplier $\delta$ for error threshold\\
\textbf{Output:} Model predictions based on $\textit{dp}$\\
\begin{algorithmic}[1] %[1] enables line numbers
\State ${M}^* \gets \text{FineTuneModel}(X_{\text{train}})$
\State $y\_preds \gets \text{EvaluateModel}(M, X_{\text{val}})$
\State $ err \gets \text{MSE}(y\_preds, y\_actual)$
\State $ \tau \gets (err + \delta * std)$
 \For{each value $ x \in X_{\text{val} }$}
\If{ \( err > \tau \) }
    \State \( X_{\text{val\_small}} \gets x \)
\EndIf
\EndFor
\State $ S_k \gets \text{LearnShapelets}(X_{\text{val\_small}})$
\State \( D_{\text{mat}} \gets \text{DISTANCE\_COMP}(X_{\text{test}}, S_k) \)
\State \( X_{\text{selected}} \gets \text{DISCARD}(dp, X_{\text{test}}, D_{mat}) \)
\State $y\_selective \gets \text{EvaluateModel}({M}^*, X_{\text{selected}})$
\Procedure{DISCARD($dp, X_{\text{test}}, D_{mat}$)}{}
% \Procedure {DISCARD}(dp, $X_{\text{test}}$, $D_{mat}$)
    \State $\text{Indices\_to\_drop} \gets \text{argmin}(D_{mat})$
     \State \Return $X_{\text{remaining\_test}}$

\EndProcedure

\Procedure{DISTANCE\_COMP($X_{\text{test}}, S_k$)}{}
    \For{each value $ x \in X_{\text{test} }$}
        \For {each value $ s \in S_{\text{k}}$}
            \State \( D_{\text{mat}}[x, s] \gets \mathrm{znorm\_ED}(x, s) \)
        \EndFor
      \EndFor
    \State \Return \( D_{\text{mat}} \)
\EndProcedure

\end{algorithmic}
\end{algorithm}

\textbf{Selective forecasting of target domain}: Based on the user-defined drop percentage, we selectively filter the predictions of those test samples that show close proximity to the shapelets learned from the validation set.

%%%%%%%%%%%%%%%%%%%%%%%%%%%%%%%%%%%%%%%%%%%%%%
\begin{table}[]
\centering
\caption{Selected datasets}
\label{tab:tsf_datasets}
% \resizebox{\columnwidth}{!}{ % Scales the table to fit within the column width
\begin{tabular}{cccccc}
\hline
\textbf{Dataset}  & \textbf{Resolution} & \textbf{Length}  \\ \hline
ETTh1      & 1 hour            & 17420                                  \\
ETTh2            & 1 hour            & 17420                                \\
ETTm1   & 15 min           & 69680                              \\
ETTm2           & 15 min            & 69680                                   \\
Exchange Rate             & 1 day             & 7588                                    \\
Traffic      & 1 hour             & 17544                                      \\
 \hline
\end{tabular}
% }
\end{table}
%%%%%%%%%%%%%%%%%%%%%%%%%%%%%%%%%%%%%%%%%%%%%%

\section{Experimental Setup}
This section describes the datasets used, the baselines adopted to compare against our framework as well as our choice of the model and its hyperparameters.

\subsection{Baseline models}
In this paper, we use TTM as the time series foundation model owing to its lightweight architecture that makes it easier to fine-tune on resource-constrained environments. Furthermore, despite its compact model size, it has shown to outperform popular benchmarks in zero-shot forecasting \cite{b2, b26}.
For robust evaluation of our proposed approach, we compare against the following baselines:

\begin{itemize}
    \item Zero-shot model (Drop percentage - 0\%).
    \item Full-shot Fine-tuned model (Drop percentage - 0\%).
    \item Random Selection using zero-shot model (Drop percentage - 20\%).
    \item Random Selection using Full-shot Fine-tuned model (Drop percentage - 20\%).
\end{itemize}

\subsection{Datasets}

To evaluate the performance of our framework, we use six open benchmark time-series datasets. Table~\ref{tab:tsf_datasets} enlists the datasets along with the resolution or sampling frequency and the length of the dataset. The Electricity Transformer Temperature datasets: ETTh1, ETTh2, ETTm1, ETTm2 \cite{b14} contain 2 years of measurements from two electricity transformers in separate Chinese counties, each with 7 sensor features. ETTh1 and ETTh2 are collected at hourly interval whereas ETTm1 and ETTm2 are collected at every 15 minute interval. All four datasets have 7 channels. However, we only use one channel as part of our univariate forecasting setting. The Exchange Rate dataset contains daily exchange rates for 8 different currencies against USD from 1990 to 2016, with XRP/USD as the target variable for forecasting. Next, we also use Traffic dataset which records the hourly rates of road occupancy on the San Francisco Freeways using 862 sensors. We split the datasets into train, validation and test splits in the same ratio as described in the original TTM paper \cite{b2}.

\subsection{Experiment details}

We use the TTM-Base model with the context length equal to 512 and forecast length set to 96. The full-shot fine-tuned version of the model involves fine-tuning the model head while keeping the backbone parameters frozen. For each dataset, we fine-tune the model on the train split of the data to get the full-shot fine-tuned version of the model. The hyperparameters for fine-tuning namely head dropout, batch-size and learning rate are based on the validation performance as reported in \cite{b2}. We run all the experiments for three random seeds and present the average results across runs. We set the drop percentage ($dp$) to 20\% and $\delta$ to 2 to derive the error threshold for all our experiments.

\begin{table*}[]
\centering
\caption{MSE obtained using Selective Forecasting guided by shapelets}
\label{tab:results_table}
\resizebox{\textwidth}{!}{ % Scales the table to fit within the column width
\label{table:results_table}
\begin{tabular}{@{}ccccccc@{}}
\toprule
\textbf{Dataset}       & \textbf{Zero-shot(ZS)} & \textbf{Full-shot Fine-Tuned (FFT)} & \textbf{\begin{tabular}[c]{@{}c@{}}Random Selection\\ (ZS)\end{tabular}} & \textbf{\begin{tabular}[c]{@{}c@{}}Random Selection\\ (FFT)\end{tabular}} & \textbf{\begin{tabular}[c]{@{}c@{}}Selective Forecasting\\ with shapelets (using ZS)\end{tabular}} & \textbf{\begin{tabular}[c]{@{}c@{}}Selective Forecasting\\ with shapelets (using FFT)\end{tabular}} \\ \midrule
\textbf{ETTh1}         & 0.0524                 & 0.0512                              & 0.0418                                                                   & \textbf{0.0408}                                                           & 0.0439                                                                                             & 0.0443                                                                                              \\
\textbf{ETTh2}         & 0.1306                 & 0.1304                              & 0.1038                                                                   & 0.1039                                                                    & 0.1002                                                                                             & \textbf{0.0992}                                                                                     \\
\textbf{ETTm1}         & 0.0275                 & 0.0264                              & 0.0220                                                                   & 0.0221                                                                    & 0.0221                                                                                             & \textbf{0.0215}                                                                                     \\
\textbf{ETTm2}         & 0.0765                 & 0.0646                              & 0.0615                                                                   & \textbf{0.0518}                                                           & 0.0644                                                                                             & 0.0555                                                                                              \\
\textbf{Exchange rate} & 0.0790                 & 0.0776                              & 0.0626                                                                   & 0.0616                                                                    & 0.0492                                                                                             & \textbf{0.0484}                                                                                     \\
\textbf{Traffic}       & 0.1766                 & 0.1173                              & 0.1412                                                                   & 0.0936                                                                    & 0.1407                                                                                             & \textbf{0.0933}                                                                                     \\
                       &                        &                                     & \multicolumn{1}{l}{}                                                     &                                                                           &                                                                                                    &                                                                                                     \\ \bottomrule
\end{tabular}
}
\end{table*}

%%%%%%%%%%%%%%%%%%%%%%%%%%%%%%%%%%%%%%%%%%%%%
\begin{figure}[t]
    \centering
    \includegraphics[width=0.45\textwidth]{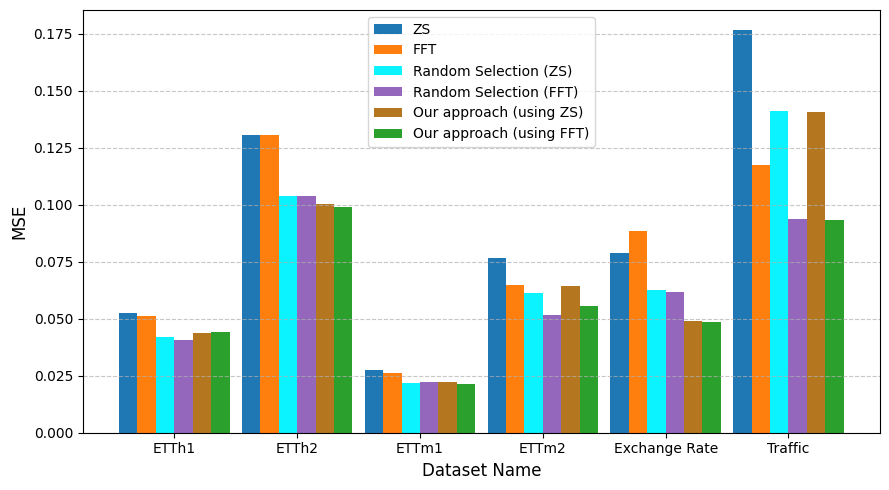}
    \caption{Performance comparison of Selective Forecasting using shapelets against baselines.}
    \label{fig:bar_plot}
\end{figure}

%%%%%%%%%%%%%%%%%%%%%%%%%%%%%%%%%%%%%%%%%%%%%

%%%%%%%%%%%%%%%%%%%%%%%%%%%%%%%%%%%%%%%%%%%%%
\begin{figure}[t]
    \centering
    \includegraphics[width=0.40\textwidth]{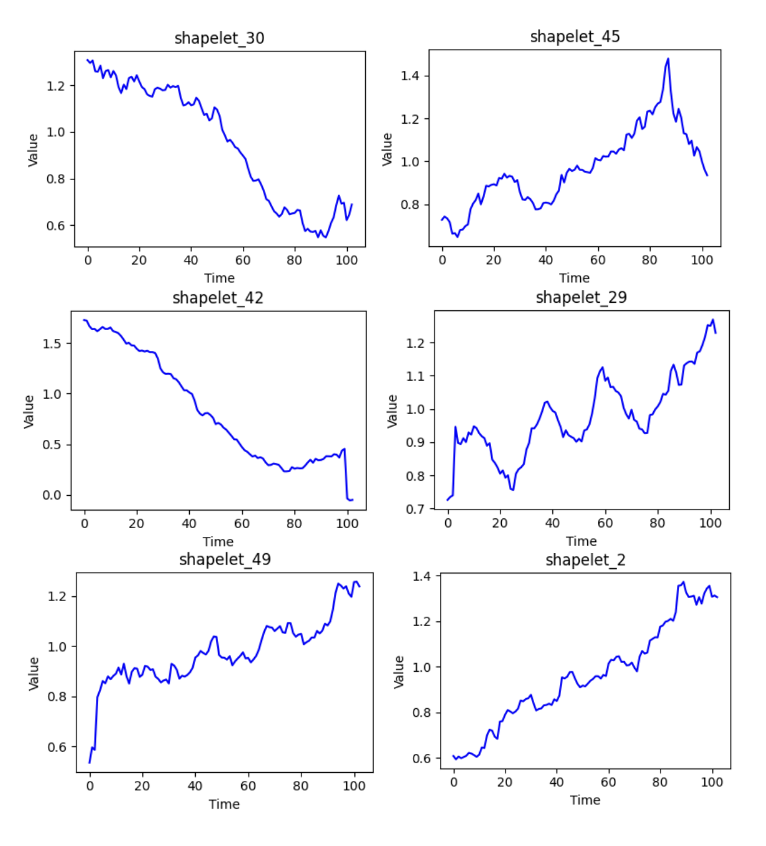}
    \caption{The plot showing a subset of shapelets learned from selected validation samples of Exchange Rate.}
    \label{fig:example_shapelets}
\end{figure}

%%%%%%%%%%%%%%%%%%%%%%%%%%%%%%%%%%%%%%%%%%%%%

\section{Results and Discussion}

In this section, we present the results obtained on running the experiments using the models and datasets discussed in the previous section. We evaluate the effectiveness of our approach by calculating the mean squared error (MSE) on the test split of the data. Table~\ref{tab:results_table} shows the average MSE obtained for each dataset. We use selective forecasting guided by shapelets to discard the test samples that are prone to high error. We use shapelets learned from the validation set to establish distance based similarity with the test samples using the distance metric: z-normalized euclidean distance. The closest matching test instances are selected to be discarded based on the user-defined drop percentage. From the MSE scores shown in Table~\ref{tab:results_table}, we observe that for four out of six datasets, our approach of selective forecasting yields lowered error improving the overall predictive performance while discarding unreliable predictions. Although, random selection using full-shot fine tuned model baseline performs slightly better for two datasets, it leads to significantly higher error for the remaining datasets. Our approach, on the other hand, provides a visual insight into the selection process and only closest matching test samples are chosen to be discarded at the inference time.

%%%%%%%%%%%%%%%%%%%%%%%%%%%%%%%%%%%%%%%%%%%%%
\begin{figure}[t]
    \centering
    \includegraphics[width=0.45\textwidth]{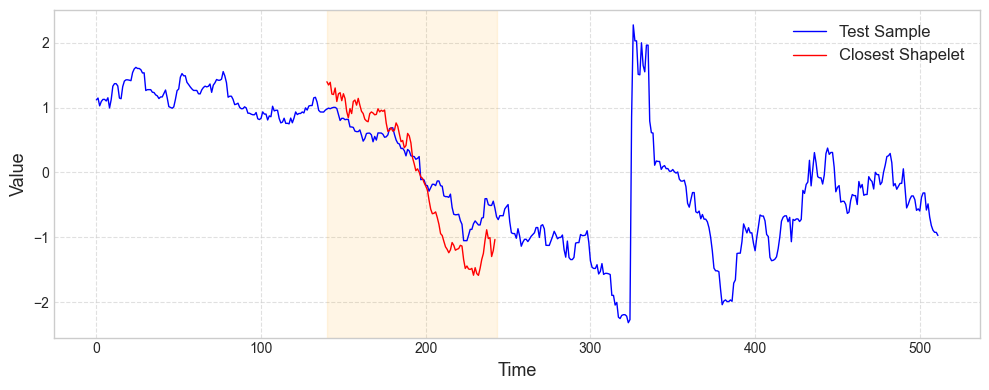}
    \caption{A test sample selected for discarding predictions with the best matching shapelet for Exchange Rate.}
    \label{fig:shapelet_match}
\end{figure}

%%%%%%%%%%%%%%%%%%%%%%%%%%%%%%%%%%%%%%%%%%%%%

%%%%%%%%%%%%%%%%%%%%%%%%%%%%%%%%%%%%%%%%%%%%%
\begin{figure}[t]
    \centering
    \includegraphics[width=0.45\textwidth]{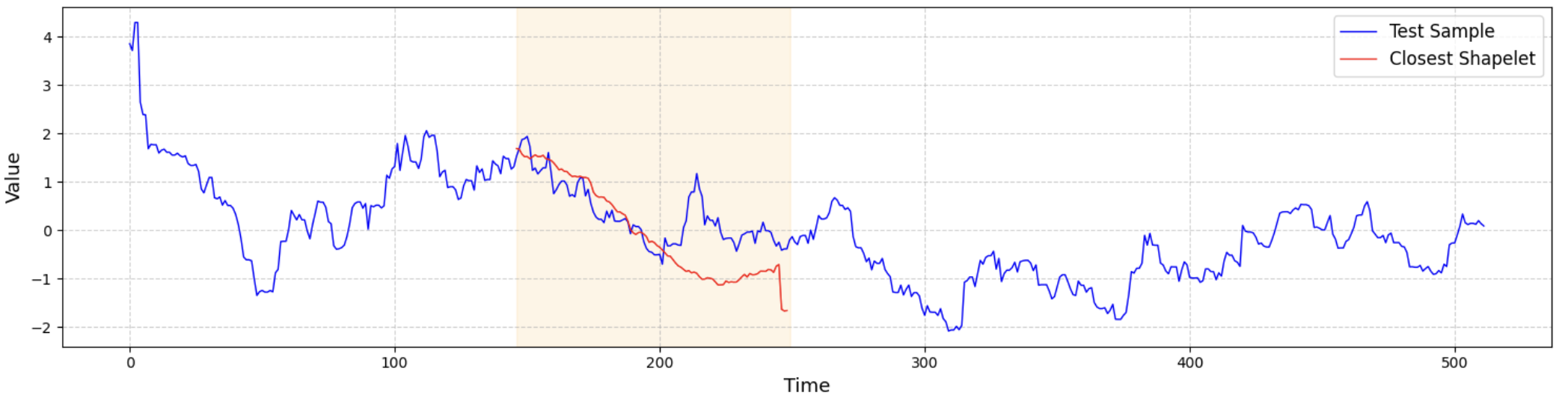}
    \caption{A test sample selected for making predictions with the best matching shapelet for Exchange Rate.}
    \label{fig:shapelet_mismatch}
\end{figure}

%%%%%%%%%%%%%%%%%%%%%%%%%%%%%%%%%%%%%%%%%%%%%

\begin{figure*}[!h]
\centering
\includegraphics[width=\textwidth]{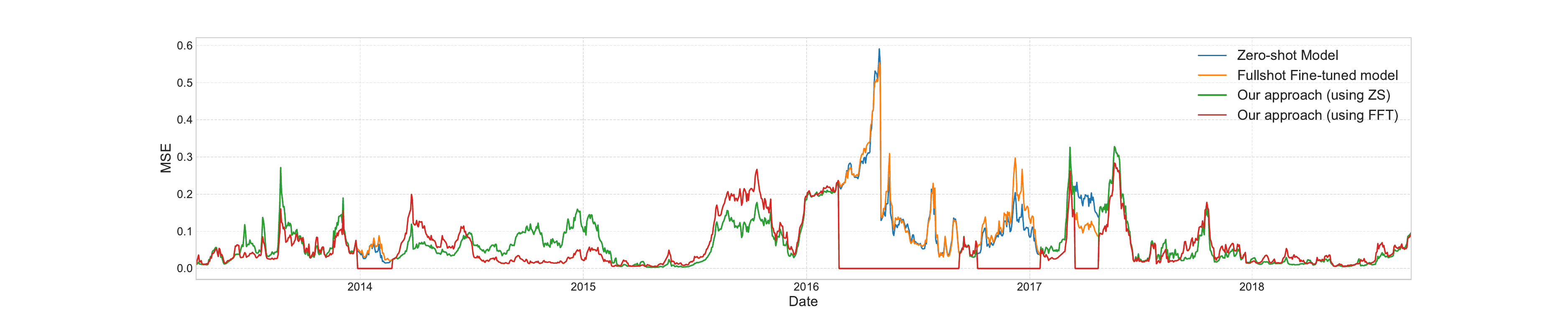}
\caption{Error comparison using our approach with the baselines for Exchange Rate.}
\label{fig:error_comparison}
\end{figure*}

Figure~\ref{fig:bar_plot} presents the results from Table~\ref{tab:results_table} in a bar plot which clearly shows green and purple bars lead to the lowest errors across all datasets. These results exhibit significant improvement in predictive performance specifically for exchange rate and traffic datasets. This shows that informed selective forecasting can significantly contribute to improved forecasting performance for TSFMs.

Figure~\ref{fig:example_shapelets} illustrates a subset of shapelets learned from the selected validation samples using Exchange Rate dataset. The mean error on validation set was 0.4506 and $\delta$ was set to 2 which resulted in the error threshold = 1.6764. These samples correspond to high prediction errors highlighting distinct patterns where the TSFM is most likely to lead to unreliable predictions. In Figure~\ref{fig:shapelet_match}, we show an example of a test sample from Exchange rate dataset that was discarded from the predictions due to a close match (z-norm euclidean distance = 4.46) with the shapelet highlighted in red. However, another test sample that was selected to make predictions is shown in Figure~\ref{fig:shapelet_mismatch}. In this case, the closest matching shapelet shows a high z-norm euclidean distance of 7.44. We also observe that this shapelet does not match the upward peak in the second half of the shaded region of the plot. This further validates our approach, which effectively utilizes shapelet-based similarity for selective forecasting.

Figure~\ref{fig:error_comparison} illustrates the average error obtained on the test set using zero-shot model, full-shot fine-tuned and our approach using both zero-shot and full-shot fine-tuned model on the Exchange rate dataset. We observe that using our approach guided by shapelets, we are able to identify test samples which are similar to the patterns captured by shapelets. These test samples correspond to higher than average error as shown by peaks during the year 2016 in figure ~\ref{fig:error_comparison}. We then selectively drop these samples at inference time such that the corresponding prediction error falls to zero as shown in the plot using the red line. Figure~\ref{fig:shapelet_match} depicts an example test sample showing a steep downward trend just before a spike which closely matches to the learned shapelet in red. Since this shapelet characterizes a pattern which is correlated to high error in the validation set, we selectively drop the predictions for this sample at test time.

\begin{table}[]
\centering
\caption{Ablation studies by varying the error threshold for filtering validation samples.}
\label{tab:ablation_results}

\label{table:ablation_results}
\begin{tabular}{@{}ccl@{}}
\toprule
\multirow{2}{*}{\textbf{\begin{tabular}[c]{@{}c@{}}Error Threshold\\ (mean error + $\delta$ * std )\end{tabular}}} & 
\multicolumn{2}{c}{\textbf{\begin{tabular}[c]{@{}c@{}}MSE using Selective\\ Forecasting\end{tabular}}} \\ 
\cmidrule{2-3}
& \multicolumn{1}{l}{Exchange Rate} & ETTh2 \\ 
\cmidrule{1-3}  
$\delta$ = 1.0 & 0.0526 & 0.1013 \\
$\delta$ = 1.5 & 0.0516 & 0.1040 \\
$\delta$ = 2.0 & 0.0484 & 0.0992 \\
$\delta$ = 2.5 & 0.0516 & 0.1028 \\
$\delta$ = 3.0 & 0.0507 & 0.1017 \\ 
\bottomrule
\end{tabular}
\end{table}

\subsection{Ablation Study}
\begin{figure}[t]
\centering
\includegraphics[width=0.45\textwidth]{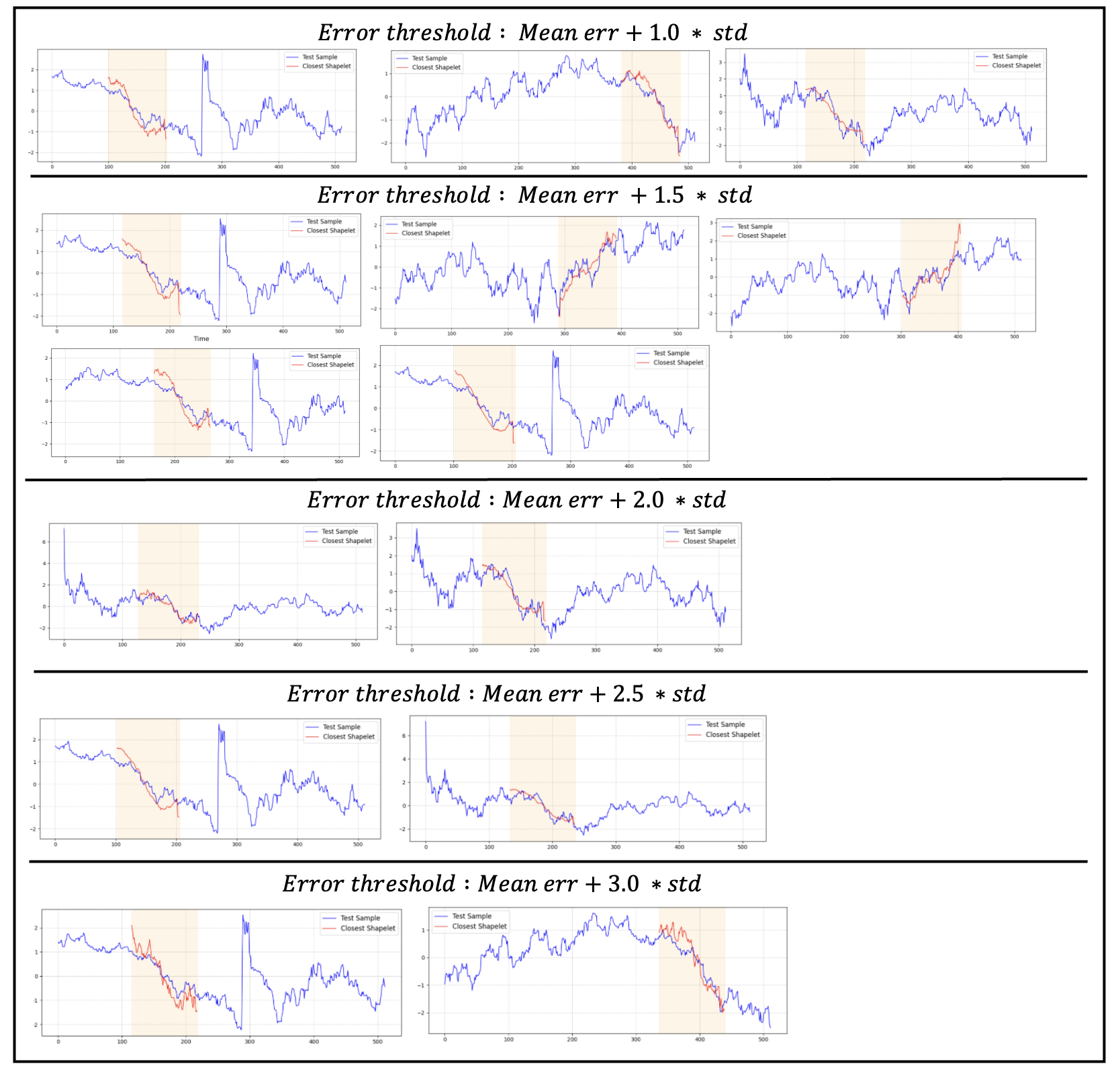}
\caption{Matching shapelets to the selected test samples for dropping predictions for different error thresholds.}
\label{fig:ablation_plot}
\end{figure}

We also performed an ablation study by varying the error threshold for filtering the validation samples used for shapelet learning for two datasets, ETTh2 and Exchange Rate. We add multiples of standard deviation to the mean error to increase the error threshold and observe the resulting shapelets learned from the validation data samples. This analysis shows that as the threshold increases, fewer validation samples are selected for shapelet learning. Figure~\ref{fig:ablation_plot} shows the unique set of shapelets and their corresponding closest matching test samples which are selected to be discarded while making predictions. We can clearly observe from the plot that, for error thresholds upto 1.5 times the standard deviation, more diverse shapelets are learned capturing both upward and downward trends. However, as the threshold increases, there are fewer validation set samples available to learn the shapelets which risks missing the patterns that might lead to high errors. This can be validated by the MSE reported in Table~\ref{tab:ablation_results}.

%%%%%%%%%%%%%%%%%%%%%%%%%%%%%%%%%%%%%%%%%%%%%%
%%%%%%%%%%%%%%%%%%%%%%%%%%%%%%%%%%%%%%%%%%%%%%

\section{Conclusions and Future Work}
Pre-trained TSFMs have shown to provide robust zero-shot downstream performance for time series forecasting tasks. However, there is immense scope of improving the average zero-shot performance by specifically focusing at time series segments where the model predictions are highly unreliable. In this paper, we presented our approach to selective forecasting in the context of TSFMs guided by using shapelets. We introduced a shapelet-based sample selection module that helped in discarding high error-prone model predictions. In order to achieve this goal, we used z-normalized euclidean distance metric to establish distance-based similarity between the learned shapelets and test samples. We demonstrate that our approach reduces the overall error by an average of 22.17\% for zero-shot and 22.62\% for full-shot fine-tuned models. It also achieves notable gains, improving the performance up to 21.41\% and 21.43\% over random selection baselines for zero-shot and full-shot fine-tuned models respectively, on one of the datasets.

\subsubsection{Acknowledgments.}
 This work has been partially supported by the 6G-XCEL project (grant agreement 101139194), funded by the EU Horizon Europe program.

\bibliography{aaai2026}

@inproceedings{b1,
  title={Learning with rejection},
  author={Cortes, Corinna and DeSalvo, Giulia and Mohri, Mehryar},
  booktitle={International conference on algorithmic learning theory},
  pages={67--82},
  year={2016},
  organization={Springer}
}

@article{b2,
  title={Tiny time mixers (ttms): Fast pre-trained models for enhanced zero/few-shot forecasting of multivariate time series},
  author={Ekambaram, Vijay and Jati, Arindam and Dayama, Pankaj and Mukherjee, Sumanta and Nguyen, Nam and Gifford, Wesley M and Reddy, Chandra and Kalagnanam, Jayant},
  journal={Advances in Neural Information Processing Systems},
  volume={37},
  pages={74147--74181},
  year={2024}
}

@inproceedings{b3,
  title={Efficient shift-invariant dictionary learning},
  author={Zheng, Guoqing and Yang, Yiming and Carbonell, Jaime},
  booktitle={Proceedings of the 22nd ACM SIGKDD international conference on knowledge discovery and data mining},
  pages={2095--2104},
  year={2016}
}

@inproceedings{b4,
  title={Time series shapelets: a new primitive for data mining},
  author={Ye, Lexiang and Keogh, Eamonn},
  booktitle={Proceedings of the 15th ACM SIGKDD international conference on Knowledge discovery and data mining},
  pages={947--956},
  year={2009}
}

@inproceedings{b5,
  title={Learning time-series shapelets},
  author={Grabocka, Josif and Schilling, Nicolas and Wistuba, Martin and Schmidt-Thieme, Lars},
  booktitle={Proceedings of the 20th ACM SIGKDD international conference on Knowledge discovery and data mining},
  pages={392--401},
  year={2014}
}

@article{b6,
  title={Time series shapelets: a novel technique that allows accurate, interpretable and fast classification},
  author={Ye, Lexiang and Keogh, Eamonn},
  journal={Data mining and knowledge discovery},
  volume={22},
  number={1},
  pages={149--182},
  year={2011},
  publisher={Springer}
}

@article{b7,
  title={Chronos: Learning the language of time series},
  author={Ansari, Abdul Fatir and Stella, Lorenzo and Turkmen, Caner and Zhang, Xiyuan and Mercado, Pedro and Shen, Huibin and Shchur, Oleksandr and Rangapuram, Syama Sundar and Arango, Sebastian Pineda and Kapoor, Shubham and others},
  journal={arXiv preprint arXiv:2403.07815},
  year={2024}
}

@article{b8,
  title={Unified training of universal time series forecasting transformers},
  author={Woo, Gerald and Liu, Chenghao and Kumar, Akshat and Xiong, Caiming and Savarese, Silvio and Sahoo, Doyen},
  journal={International Conference on Machine Learning (ICML)},
  year={2024}
}

@article{b9,
  title={A decoder-only foundation model for time-series forecasting},
  author={Das, Abhimanyu and Kong, Weihao and Sen, Rajat and Zhou, Yichen},
  journal={International Conference on Machine Learning (ICML)},
  year={2023}
}

@inproceedings{b10,
  title={Large Language Models Are Zero-Shot Time Series Forecasters},
  author={Gruver, Nate and Finzi, Marc Anton and Qiu, Shikai and Wilson, Andrew Gordon},
  booktitle={Thirty-seventh Conference on Neural Information Processing Systems},
  year={2023}
}

@inproceedings{b11,
  title={{One Fits All}: Power General Time Series Analysis by Pretrained LM},
  author={Zhou, Tian and Niu, Peisong and Wang, Xue and Sun, Liang and Jin, Rong},
  booktitle={NeurIPS},
  year={2023}
}

@article{b12,
  title={Autoformer: Decomposition transformers with auto-correlation for long-term series forecasting},
  author={Wu, Haixu and Xu, Jiehui and Wang, Jianmin and Long, Mingsheng},
  journal={Advances in neural information processing systems},
  volume={34},
  pages={22419--22430},
  year={2021}
}

@inproceedings{b13,
  title={Fedformer: Frequency enhanced decomposed transformer for long-term series forecasting},
  author={Zhou, Tian and Ma, Ziqing and Wen, Qingsong and Wang, Xue and Sun, Liang and Jin, Rong},
  booktitle={International conference on machine learning},
  pages={27268--27286},
  year={2022},
  organization={PMLR}
}

@inproceedings{b14,
  title={Informer: Beyond efficient transformer for long sequence time-series forecasting},
  author={Zhou, Haoyi and Zhang, Shanghang and Peng, Jieqi and Zhang, Shuai and Li, Jianxin and Xiong, Hui and Zhang, Wancai},
  booktitle={Proceedings of the AAAI conference on artificial intelligence},
  volume={35},
  pages={11106--11115},
  year={2021}
}

@article{b15,
  title={Timesnet: Temporal 2d-variation modeling for general time series analysis},
  author={Wu, Haixu and Hu, Tengge and Liu, Yong and Zhou, Hang and Wang, Jianmin and Long, Mingsheng},
  journal={arXiv preprint arXiv:2210.02186},
  year={2022}
}

@article{b16,
  title={Agnostic pointwise-competitive selective classification},
  author={Wiener, Yair and El-Yaniv, Ran},
  journal={Journal of Artificial Intelligence Research},
  volume={52},
  pages={171--201},
  year={2015}
}

@article{b17,
  title={Boosting with abstention},
  author={Cortes, Corinna and DeSalvo, Giulia and Mohri, Mehryar},
  journal={Advances in neural information processing systems},
  volume={29},
  year={2016}
}

@article{b18,
  title={Selective classification for deep neural networks},
  author={Geifman, Yonatan and El-Yaniv, Ran},
  journal={Advances in neural information processing systems},
  volume={30},
  year={2017}
}

@article{b20,
  title={An optimum character recognition system using decision functions},
  author={Chow, Chi-Keung},
  journal={IRE Transactions on Electronic Computers},
  number={4},
  pages={247--254},
  year={2009},
  publisher={IEEE}
}

@article{b21,
  title={On the Foundations of Noise-free Selective Classification.},
  author={El-Yaniv, Ran and others},
  journal={Journal of Machine Learning Research},
  volume={11},
  number={5},
  year={2010}
}

@article{b22,
  title={Lora: Low-rank adaptation of large language models},
  author={Hu, Edward J and Shen, Yelong and Wallis, Phillip and Allen-Zhu, Zeyuan and Li, Yuanzhi and Wang, Shean and Wang, Lu and Chen, Weizhu},
  journal={arXiv preprint arXiv:2106.09685},
  year={2021}
}

@article{b23,
  title={Parameter Efficient Fine Tuning: A Comprehensive Analysis Across Applications},
  author={Balne, Charith Chandra Sai and Bhaduri, Sreyoshi and Roy, Tamoghna and Jain, Vinija and Chadha, Aman},
  journal={arXiv preprint arXiv:2404.13506},
  year={2024}
}

@article{b24,
  title={Multi-Scale Finetuning for Encoder-based Time Series Foundation Models},
  author={Qiao, Zhongzheng and Liu, Chenghao and Zhang, Yiming and Jin, Ming and Pham, Quang and Wen, Qingsong and Suganthan, PN and Jiang, Xudong and Ramasamy, Savitha},
  journal={arXiv preprint arXiv:2506.14087},
  year={2025}
}

@article{b25,
  title={Less is More: Unlocking Specialization of Time Series Foundation Models via Structured Pruning},
  author={Zhao, Lifan and Shen, Yanyan and Liu, Zhaoyang and Wang, Xue and Deng, Jiaji},
  journal={arXiv preprint arXiv:2505.23195},
  year={2025}
}

@article{b26,
  title={A survey of time series foundation models: Generalizing time series representation with large language model},
  author={Ye, Jiexia and Zhang, Weiqi and Yi, Ke and Yu, Yongzi and Li, Ziyue and Li, Jia and Tsung, Fugee},
  journal={arXiv preprint arXiv:2405.02358},
  year={2024}
}

@article{b27,
  title={AT4TS: Autotune for Time Series Foundation Models},
  author={Tomar, Shivani and Tirupathi, Seshu and Marinescu, Radu and Daly, Elizabeth M and Dusparic, Ivana},
  journal={Transactions on Machine Learning Research},
    year={2025}
}

@article{b28,
  title={Foundts: Comprehensive and unified benchmarking of foundation models for time series forecasting},
  author={Qiu, Xiangfei and Chen, Peng and Wang, Yihang and Cheng, Hanyin and Shu, Yang and Hu, Jilin and Guo, Chenjuan and Zhou, Aoying and Wen, Qingsong and Jensen, Christian S and others},
journal={arXiv preprint arXiv:2410.11802v1},
  year={2024}
}

\section{Appendix}

We conducted additional experiments including ablation studies by varying the drop percentage (dp) parameter. The results are reported in Table~\ref{tab:ablation_results2}. The error threshold $\delta$ was set to 2 same as the main experiments. We can confirm that as the drop percentage increases for each dataset, the MSE decreases monotonically.

\begin{table}[!h]
\caption{Ablation studies by varying the drop percentage (dp) in the range of \{10\%, 20\%, 30\%, 40\%, 50\%\} using our approach.} 
\label{tab:ablation_results2}
\resizebox{\columnwidth}{!}{ % Scales the table to fit within the column width
\label{table:ablation_results2}
\begin{tabular}{cccccc}
\hline
\multirow{2}{*}{\textbf{Dataset}} & \multicolumn{5}{c}{\textbf{\begin{tabular}[c]{@{}c@{}}Selective Forecasting\\ with shapelets (using FFT)\end{tabular}}}              \\
                                  & \multicolumn{1}{l}{10\%} & \multicolumn{1}{l}{20\%} & \multicolumn{1}{l}{30\%} & \multicolumn{1}{l}{40\%} & \multicolumn{1}{l}{50\%} \\ \hline
\textbf{ETTh1}                    & 0.0482                   & 0.0442                   & 0.0400                   & 0.0344                   & 0.0312                   \\
\textbf{ETTh2}                    & 0.1126                   & 0.0992                   & 0.0872                   & 0.0758                   & 0.0654                   \\
\textbf{ETTm1}                    & 0.0253                   & 0.0215                   & 0.0195                   & 0.0165                   & 0.0133                   \\
\textbf{ETTm2}                    & 0.0607                   & 0.0555                   & 0.0484                       & 0.0413                   & 0.0341                   \\
\textbf{Exchange rate}            & 0.0622                   & 0.0484                   & 0.0375                   & 0.0293                   & 0.0266                   \\
\textbf{Traffic}                  & 0.1031                   & 0.0933                   & 0.0787                   & 0.0669                   & 0.0567                   \\
                                  &                          &                          & \multicolumn{1}{l}{}     &                          &                          \\ \hline
\end{tabular}
}
\end{table}

\begin{figure}[!h]
    \centering
    \includegraphics[width=0.45\textwidth]{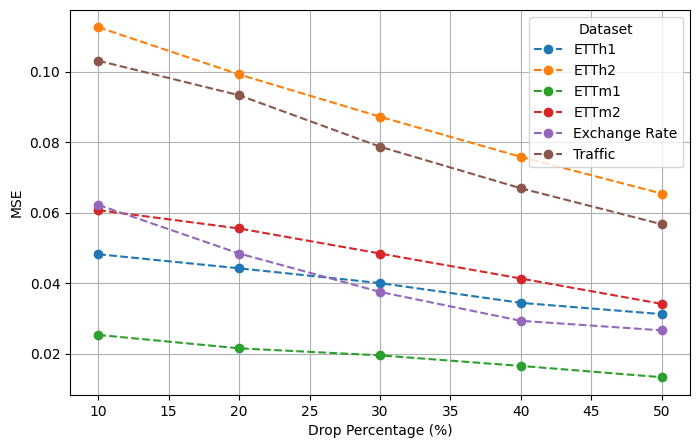}
    \caption{Plot showing the change in MSE as the drop percentage increases using Selective Forecasting (using FFT) for all datasets.}
    \label{fig:ablation_dp}
\end{figure}

Figure~\ref{fig:ablation_dp} shows the error plot as we increase the drop percentage for each dataset. The reported results correspond to our approach, selective forecasting using full-shot fine-tuned model. We observe that the monotonically decreasing error trend is consistent across all datasets. 

%%%%%%%%%%%%%%%%%%%%%%%%%%%%%%%%%%%%%%%%%%%%%

%%%%%%%%%%%%%%%%%%%%%%%%%%%%%%%%%%%%%%%%%%%%%

\begin{figure}[]
    \centering
    \includegraphics[width=0.45\textwidth]{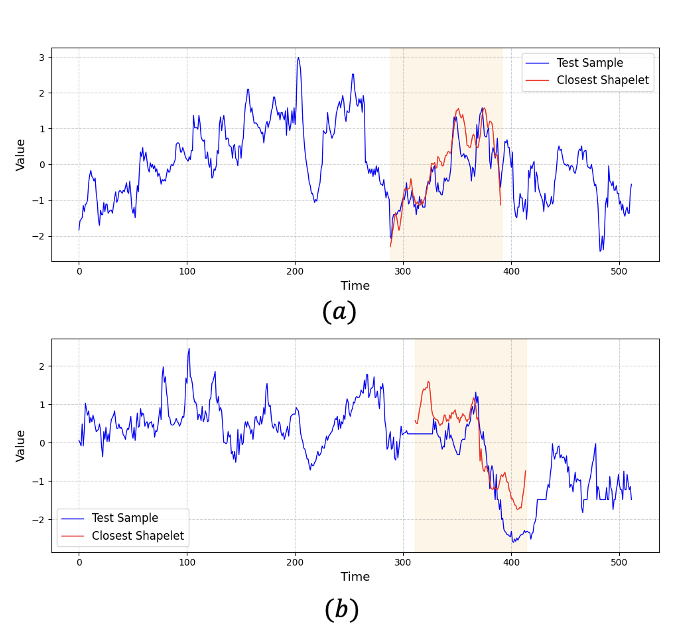}
    \caption{The figure (a) shows a discarded sample with the best closest match shapelet (d=5.47) and (b) shows a sample selected for prediction with the worst closest match shapelet (d=7.86) for ETTh1.}
    \label{fig:bw_etth1}
\end{figure}

%%%%%%%%%%%%%%%%%%%%%%%%%%%%%%%%%%%%%%%%%%%%%
\begin{figure}[]
    \centering
    \includegraphics[width=0.45\textwidth]{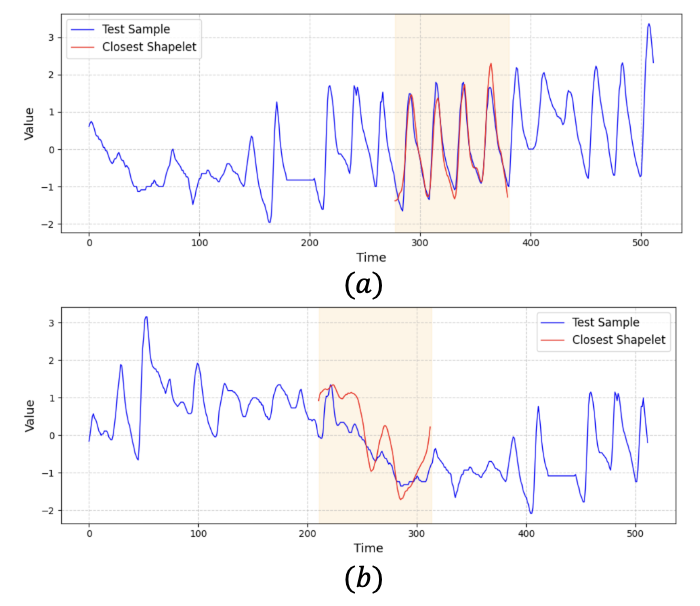}
    \caption{The figure (a) shows a discarded sample with the best closest match shapelet (d=2.78) and (b) shows a sample selected for prediction with the worst closest match shapelet (d=6.16) for ETTh2.}
    \label{fig:bw_etth2}
\end{figure}

%%%%%%%%%%%%%%%%%%%%%%%%%%%%%%%%%%%%%%%%%%%%%

\begin{figure*}[!h]
\centering
\includegraphics[width=0.9\textwidth]{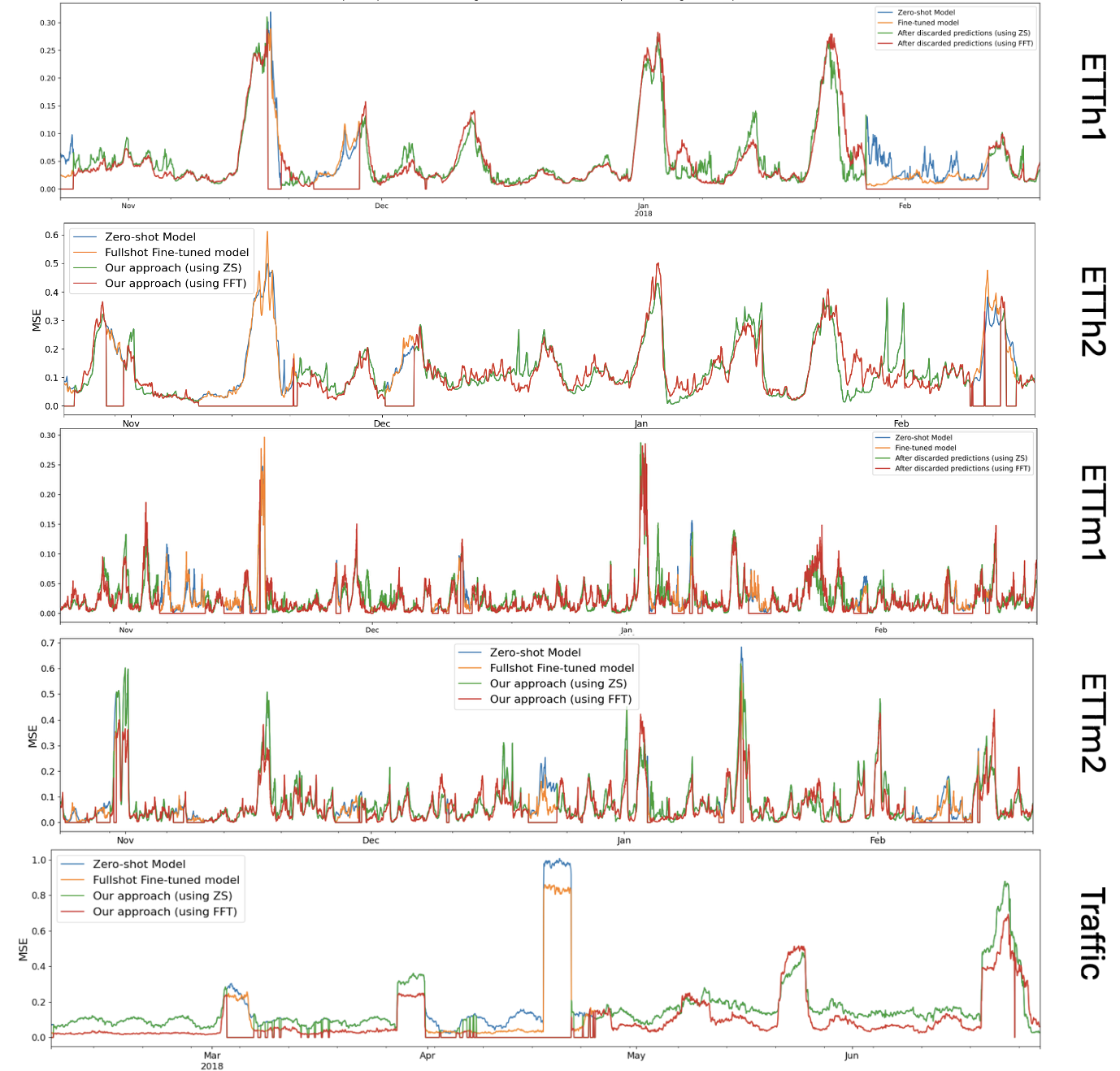}
\caption{Error comparison using our approach with the baselines for the remaining datasets.}
\label{fig:error_all_datasets}
\end{figure*}

%%%%%%%%%%%%%%%%%%%%%%%%%%%%%%%%%%%%%%%%%%%%%

In the following sections, we also illustrate test samples for each dataset showing the best and the worst matching shapelets and the corresponding error curve after using selective forecasting. Figure~\ref{fig:bw_etth1} clearly depicts how shapelet based similarity assists in the selective forecasting process for ETTh1 dataset. The error plot for ETTh1 in figure~\ref{fig:error_all_datasets} shows that the shapelet matching helps identify samples leading to high prediction errors for the first two peaks. However, we also observe that test samples leading to high error peaks at the start and end of January 2018 do not get discarded. This is highly likely due to distribution shift between the validation and test set. In such cases, shapelets learned on validation set are unable to represent the changes in the patterns/trends that may arise later. Future work will explore expanding the approach to encompass adaptive shapelet learning, further strengthening its applicability across scenarios.

\begin{figure}[]
    \centering
    \includegraphics[width=0.45\textwidth]{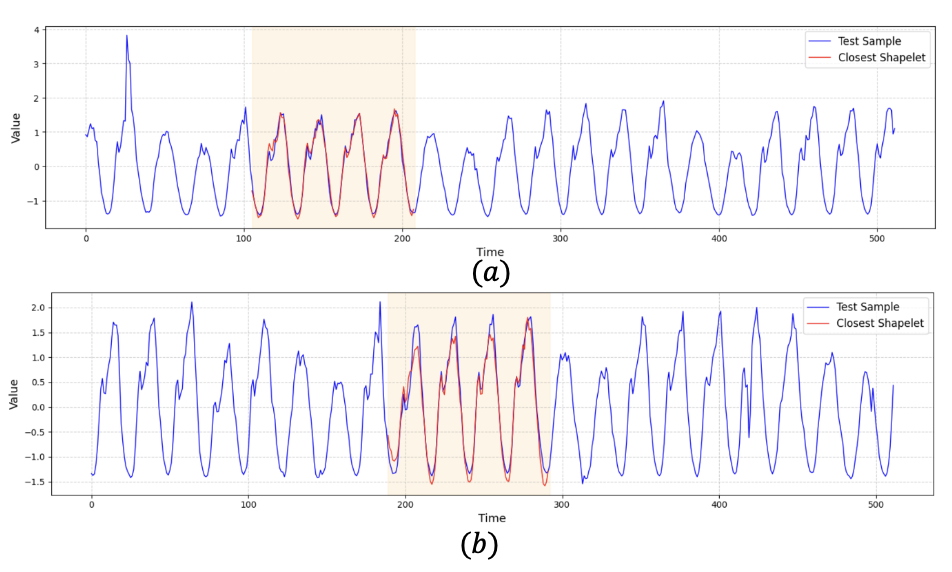}
    \caption{The figure (a) shows a discarded sample with the best closest match shapelet (d=1.43) and (b) shows a sample selected for prediction with the worst closest match shapelet (d=2.18) for Traffic.}
    \label{fig:bw_traffic}
\end{figure}

\begin{figure}[]
    \centering
    \includegraphics[width=0.45\textwidth]{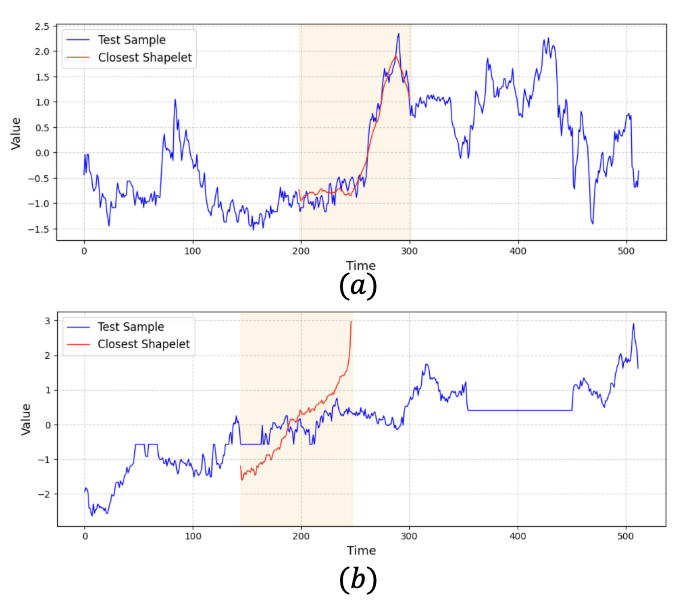}
    \caption{The figure (a) shows a discarded sample with the best closest match shapelet (d=2.23) and (b) shows a sample selected for prediction with the worst closest match shapelet (d=7.67) for ETTm1.}
    \label{fig:bw_ettm1}
\end{figure}

\begin{figure}[]
    \centering
    \includegraphics[width=0.45\textwidth]{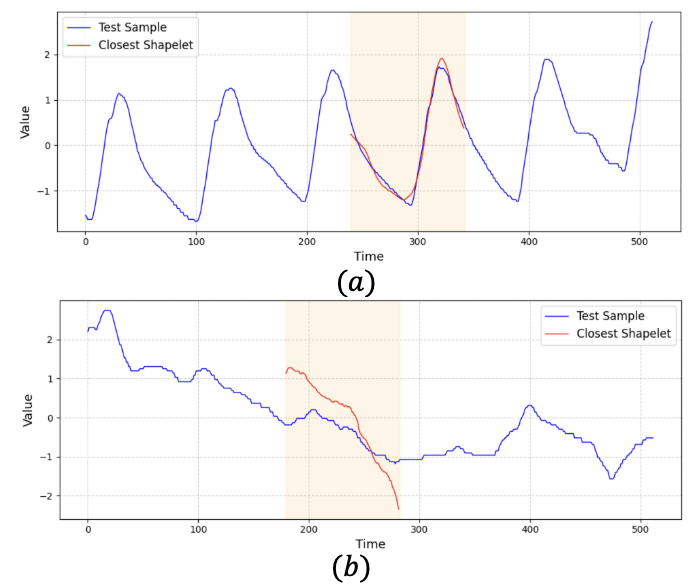}
    \caption{The figure (a) shows a discarded sample with the best closest match shapelet (d=1.27) and (b) shows a sample selected for prediction with the worst closest match shapelet (d=7.38) for ETTm2.}
    \label{fig:bw_ettm2}
\end{figure}

\end{document}